\documentclass{article}

\usepackage{amsmath}
\usepackage{hyperref}

\title{\texttt{word2vec} Explained: Deriving Mikolov et al.'s\\Negative-Sampling
Word-Embedding Method}
\author{Yoav Goldberg and Omer Levy\\\texttt{\{yoav.goldberg,omerlevy\}@gmail.com}}
\date{February 14, 2014}
\begin{document}
\maketitle

The \texttt{word2vec} software of Tomas Mikolov and
colleagues\footnote{\url{https://code.google.com/p/word2vec/}} has gained a lot
of traction lately, and provides state-of-the-art word embeddings.
The learning models behind the software are described in two research papers
\cite{mikolov2013efficient,mikolov2013distributed}.
We found the description of the models in these papers to be somewhat cryptic
and hard to follow.  While the motivations and presentation
may be obvious to the neural-networks language-modeling crowd, we had to struggle
quite a bit to figure out the rationale behind the equations.

This note is an attempt to explain equation (4) (\textit{negative sampling}) 
in ``Distributed Representations of
Words and Phrases and their Compositionality'' by Tomas Mikolov, Ilya Sutskever,
Kai Chen, Greg Corrado and Jeffrey Dean \cite{mikolov2013distributed}.

\section{The skip-gram model}

The departure point of the paper is the skip-gram model. In this model we are
given a corpus of words $w$ and their contexts $c$. We consider the conditional
probabilities $p(c|w)$, and given a corpus $Text$, the goal is to set the
parameters $\theta$ of $p(c|w;\theta)$ so as to maximize the corpus probability:

\begin{equation}
\arg\max_\theta \prod_{w\in Text}\left[\prod_{c \in C(w)} p(c|w;\theta)\right]
\end{equation}

\noindent in this equation, $C(w)$ is the set of contexts of word $w$.
Alternatively:

\begin{equation}
\arg\max_\theta \prod_{(w,c)\in D} p(c|w;\theta)
\label{eq:obj1}
\end{equation}

\noindent here $D$ is the set of all word and context pairs we extract from the
text.

\subsection{Parameterization of the skip-gram model}

\noindent One approach for parameterizing the skip-gram model follows the
neural-network language models literature, and models the conditional probability
$p(c|w;\theta)$ using soft-max:

\begin{equation}
p(c|w;\theta) = \frac{e^{v_c \cdot v_w}}{\sum_{c'\in C}e^{v_{c'} \cdot v_w}}
\end{equation}

\noindent where $v_c$ and $v_w \in R^d$ are vector representations for $c$ and $w$
respectively, and $C$ is the set of all available contexts.\footnote{Throughout
this note, we assume that the words and the contexts come from distinct
vocabularies, so that, for example, the vector associated with the word
\emph{dog} will be different from the vector associated with the context
\emph{dog}.  This assumption follows the literature, where it is not
motivated.  One motivation for making this assumption is the following: consider
the case where both the word \emph{dog} and the context \emph{dog} share the
same vector $v$. Words
hardly appear in the contexts of themselves, and so the model should assign a
low probability to $p(dog|dog)$, which entails assigning a low value to
$v\cdot v$ which is impossible.} 
The parameters $\theta$ are $v_{c_i}$, $v_{w_i}$ for $w\in V$,
$c\in C$, $i\in 1,\cdots,d$ (a total of $|C| \times |V| \times d$ parameters).
We would like to set the parameters such that the product (\ref{eq:obj1}) is maximized.

Now will be a good time to take the log and switch from product to sum:

\begin{equation}
\arg\max_\theta \sum_{(w,c)\in D} \log p(c|w) = \sum_{(w,c)\in D} ( \log e^{v_c
\cdot v_w} - \log\sum_{c'}e^{v_{c'} \cdot v_w} )
\label{eq:obj1a}
\end{equation}

\noindent An assumption underlying the embedding process is the following:

\begin{description}
   \item[Assumption] maximizing objective \ref{eq:obj1a} will result in good
      embeddings $v_w \;\;\;\forall~w\in~V$, in the sense that similar words will
      have similar vectors.
\end{description}

\noindent It is not clear to us at this point why this assumption holds.

While objective (\ref{eq:obj1a}) can be computed, it is computationally expensive to do so,
because the term $p(c|w;\theta)$ is very expensive to compute due to the summation
$\sum_{c'\in C}e^{v_{c'} \cdot v_w}$ over all the contexts $c'$ (there can be
hundreds of thousands of them).  One way of making the computation more
tractable is to replace the softmax with an \emph{hierarchical softmax}.  We
will not elaborate on this direction. 

\section{Negative Sampling}

Mikolov et al. \cite{mikolov2013distributed} present the negative-sampling approach as a more
efficient way of deriving word embeddings.  While negative-sampling is based on
the skip-gram model, it is in fact optimizing a different objective. What
follows is the derivation of the negative-sampling objective.

Consider a pair $(w,c)$ of word and context.  Did this pair come from the
training data?
Let's denote by $p(D=1|w,c)$ the probability that $(w,c)$ came from the corpus data.
Correspondingly, $p(D=0|w,c) = 1-p(D=1|w,c)$ will be the probability that $(w,c)$ did not come
from the corpus data.
As before, assume there are parameters $\theta$ controlling the distribution:
$p(D=1|w,c;\theta)$.  Our goal is now to find parameters to maximize the
probabilities that all of the observations indeed came from the data:

\begin{align*}
& \arg\max_\theta \prod_{(w,c)\in D} p(D=1|w,c;\theta) \\
= & \arg\max_\theta \; \log\prod_{(w,c)\in D} p(D=1|w,c;\theta) \\
= & \arg\max_\theta \sum_{(w,c)\in D} \log p(D=1|w,c;\theta)
\end{align*}

\noindent
The quantity $p(D=1|c,w;\theta)$ can be defined using softmax:
\[
p(D=1|w,c;\theta) = \frac{1}{1+e^{-v_c \cdot v_w}}
\]

\noindent Leading to the objective:
\begin{align*}
\arg\max_\theta \sum_{(w,c)\in D} \log \frac{1}{1+e^{-v_c \cdot v_w}}
\end{align*}

\noindent This objective has a trivial solution if we set $\theta$ such that
$p(D=1|w,c;\theta) = 1$ for every pair $(w,c)$.  This can be easily achieved by setting
$\theta$ such that $v_c = v_w$ and  $v_c \cdot v_w = K$ for all $v_c, v_w$,
where $K$ is large enough number (practically, we get a probability of 1 as
soon as $K \approx 40$). 

We need a mechanism that prevents all the vectors from having the same value,
by disallowing some $(w,c)$ combinations. One way to do so, is to present the
model with some $(w,c)$ pairs for which
$p(D=1|w,c;\theta)$ must be low, i.e. pairs which are not in the data.
This is achieved by generating the set $D'$ of random $(w,c)$ pairs, assuming they
are all incorrect (the name ``negative-sampling'' stems from the set $D'$ of
randomly sampled negative examples).
The optimization objective now becomes:

\begin{align*}
& \arg\max_\theta \prod_{(w,c)\in D} p(D=1|c,w;\theta) \prod_{(w,c)\in D'} p(D=0|c,w;\theta) \\
= & \arg\max_\theta \prod_{(w,c)\in D} p(D=1|c,w;\theta) \prod_{(w,c)\in D'} (1-p(D=1|c,w;\theta)) \\
= & \arg\max_\theta \sum_{(w,c)\in D} \log p(D=1|c,w;\theta) + \sum_{(w,c)\in
D'} \log (1-p(D=1|w,c;\theta)) \\
= & \arg\max_\theta \sum_{(w,c)\in D} \log \frac{1}{1+e^{- v_c \cdot v_w}} +
\sum_{(w,c) \in D'} \log (1-\frac{1}{1+e^{- v_c \cdot v_w}}) \\
= & \arg\max_\theta \sum_{(w,c)\in D} \log \frac{1}{1+e^{- v_c \cdot v_w}} +
\sum_{(w,c) \in D'} \log (\frac{1}{1+e^{v_c \cdot v_w}}) \\
\end{align*}

\noindent
If we let $\sigma(x) = \frac{1}{1+e^{-x}}$ we get:
\begin{align*}
& \arg\max_\theta \sum_{(w,c)\in D} \log \frac{1}{1+e^{- v_c \cdot v_w}} +
\sum_{(w,c) \in D'} \log (\frac{1}{1+e^{v_c \cdot v_w}}) \\
= & \arg\max_\theta \sum_{(w,c)\in D} \log \sigma(v_c \cdot v_w) +
\sum_{(w,c) \in D'} \log \sigma(-v_c \cdot v_w) \\
\end{align*}

\noindent
which is almost equation (4) in Mikolov et al (\cite{mikolov2013distributed}).

The difference from Mikolov et al. is that here we present the objective for the
entire corpus $D \cup D'$, while they present it for one example $(w,c)\in D$
and $k$ examples $(w,c_j)\in D'$, following a particular way of constructing
$D'$.

Specifically, with negative sampling of $k$, Mikolov et al.'s constructed $D'$ is $k$ times
larger than $D$, and for each $(w,c) \in D$ we construct $k$ samples
$(w,c_1),\ldots,(w,c_k)$,
where each $c_j$ is drawn according to its unigram distribution raised to the
$3/4$ power. This is equivalent to drawing the samples $(w,c)$ in $D'$ from the
distribution
$(w,c)~\sim~p_{words}(w)\frac{p_{contexts}(c)^{3/4}}{Z}$, where $p_{words}(w)$ and
$p_{contexts}(c)$ are the unigram
distributions of words and contexts respectively, and $Z$ is a normalization
constant. In the work of Mikolov et al. each context is a word (and all words appear as contexts), and so
$p_{context}(x) = p_{words}(x) = \frac{count(x)}{|Text|}$

\subsection{Remarks}
\begin{itemize}
   \item Unlike the Skip-gram model described above, the formulation in this
section does not model $p(c|w)$ but instead models a quantity related to the joint
distribution of $w$ and $c$.

   \item If we fix the words representation and learn only the contexts
      representation, or fix the contexts representation and learn only the word
      representations, the model reduces to logistic regression, and is convex.
      However, in this model the words and contexts representations are learned
      jointly, making the model non-convex.
\end{itemize}

\section{Context definitions}

This section lists some peculiarities of the contexts used in the
\texttt{word2vec} software, as reflected in the code.  Generally speaking, for a
sentence of $n$ words $w_1,\dots,w_n$,
contexts of a word $w_i$ comes from a window of size $k$ around the word: $C(w)
= w_{i-k},\dots,w_{i-1},w_{i+1},\dots,w_{i+k}$, where $k$ is a parameter.
However, there are two subtleties:

\begin{description}
   \item[Dynamic window size] the window size that is being used is dynamic --
      the parameter $k$ denotes the $maximal$ window size. For
      each word in the corpus, a window size $k'$ is sampled uniformly from
      $1,\dots,k$.

   \item[Effect of subsampling and rare-word pruning] \texttt{word2vec} has
      two additional parameters for discarding some of the input words: words
      appearing less than \texttt{min-count} times are not considered as either
      words or contexts, an in addition frequent words (as defined by the
      \texttt{sample} parameter) are down-sampled.  Importantly, these words are
      removed from the text \emph{before} generating the contexts.  This has the
      effect of \emph{increasing the effective window size} for certain words. 
      According to Mikolov et al. \cite{mikolov2013distributed}, sub-sampling of frequent words improves the quality of the resulting embedding on some benchmarks.  The
      original motivation for sub-sampling was that frequent words are less
      informative.  Here we see another explanation for its effectiveness:
      the effective window size grows, including context-words which are both
      content-full and linearly far away from the focus word, thus making the
      similarities more topical.
\end{description}

\section{Why does this produce good word representations?}

Good question. We don't really know. 

The distributional hypothesis states that words in similar contexts have similar
meanings.  The objective above clearly tries to increase the quantity $v_w\cdot~
v_c$ for good word-context pairs, and decrease it for bad ones. Intuitively,
this means that words that share many contexts will be similar to each other
(note also that contexts sharing many words will also be similar to each
other).  This is, however, very hand-wavy.

Can we make this intuition more precise? We'd really like to see something more
formal.

\bibliographystyle{plain}
\bibliography{embed}

\end{document}